\pdfoutput=1

\documentclass[11pt]{article}
\usepackage[final]{acl}

\usepackage{times}
\usepackage{latexsym}

\usepackage[T1]{fontenc}

\usepackage[utf8]{inputenc}

\usepackage{microtype}

\usepackage{inconsolata}

\usepackage{graphicx}

\usepackage{color, xcolor}
\usepackage{booktabs}
\usepackage{multicol}
\usepackage{multirow, makecell, caption}
\usepackage{colortbl}
\usepackage{tikz}
\usepackage{arydshln}
\usepackage{pgfplots}
\usepackage{amsmath}
\usepackage{amssymb}
\usepackage{pgfplots}
\pgfplotsset{compat=1.18}
\usepackage{tabularx}
\usepackage{enumerate}

\definecolor{darkgreen}{RGB}{0, 100, 0}

\usepackage{listings}
\usepackage{xcolor}

\title{Beyond Correctness: Confidence-Aware Reward Modeling for \\ Enhancing Large Language Model Reasoning}

\author{
    Qianxi He\textsuperscript{1,2}, Qingyu Ren\textsuperscript{2}, Shanzhe Lei\textsuperscript{1}, Xuhong Wang\textsuperscript{1\textdagger}, Yingchun Wang
\textsuperscript{1} \\
    \textsuperscript{1}Shanghai Artificial Intelligence Laboratory \\
    \textsuperscript{2}Shanghai Key Laboratory of Data Science, School of Computer Science, Fudan University\\
    \{qxhe23, qyren24\}@m.fudan.edu.cn, \{leishanzhe, wangxuhong, wangyingchun\}@pjlab.org.cn\\
}

\begin{document}
\maketitle
\begin{abstract}

Recent advancements in large language models (LLMs) have shifted the post-training paradigm from traditional instruction tuning and human preference alignment toward reinforcement learning (RL) focused on reasoning capabilities. However, numerous technical reports indicate that purely rule-based reward RL frequently results in poor-quality reasoning chains or inconsistencies between reasoning processes and final answers, particularly when the base model is of smaller scale. During the RL exploration process, models might employ low-quality reasoning chains due to the lack of knowledge, occasionally producing correct answers randomly and receiving rewards based on established rule-based judges.
This constrains the potential for resource-limited organizations to conduct direct reinforcement learning training on smaller-scale models.
We propose a novel confidence-based reward model tailored for enhancing STEM reasoning capabilities. Unlike conventional approaches, our model penalizes not only incorrect answers but also low-confidence correct responses, thereby promoting more robust and logically consistent reasoning. We validate the effectiveness of our approach through static evaluations, Best-of-N inference tests, and PPO-based RL training. Our method outperforms several state-of-the-art open-source reward models across diverse STEM benchmarks. We release our codes and model in \url{https://github.com/qianxiHe147/C2RM}.
\end{abstract}

\renewcommand{\thefootnote}{\textdagger}
\footnotetext{Corresponding author.}

\section{Introduction}

In the past, the traditional post-training process of large language models (LLMs) usually included instructional fine-tuning and human preference alignment~\cite{kaddour2023challenges,christiano2017deep,bai2022training,wang2024interpretable}. 
Recently, reasoning models such as OpenAI o1~\cite{jaech2024openai} and Deepseek R1~\cite{guo2025deepseek} have transformed the LLMs post-training paradigm to lightweight chain-of-thought (CoT)~\cite{wei2022chain} bootstrapping combined with rule-based STEM reinforcement learning (RL). 


However, numerous technical reports~\cite{ding2024break, wei2022chain} indicate that purely rule-based reward RL frequently results in poor-quality reasoning chains or inconsistencies between reasoning processes and final answers, particularly when the base model is of smaller scale. The rule-based approach, while effective for evaluating final answers, provides insufficient guidance for optimizing the intermediate reasoning steps that lead to those answers, creating a disconnect between the reward signal and the desired reasoning behavior.


To confront this challenge, open-source LLMs including Deepseek-R1, Qwen3~\cite{yang2025qwen3}, and Llama4 choose to implement RL on foundation models with hundreds of billions of parameters. Large-scale models demonstrate superior capacity to maintain robust cognitive processes during RL regimens. Subsequently, the more compact variants are distilled from their immensely scaled foundational counterparts. This paradigmatic constraint proves exceedingly inhospitable to resource-constrained organizations, effectively extinguishing their prospects of harnessing RL to enhance performance on domain-specific endeavors.


We believe that the fundamental reason why smaller-scale models struggle to make progress in RL lies in their insufficient internal world knowledge to generate high-confidence responses for certain challenging problems. During the RL exploration process, the models might employ low-quality or logic inconsistent reasoning chains, occasionally and randomly producing correct answers and receiving rewards based on established rule-based judge. The key to addressing this issue is whether we can penalize the model's "low-confidence" responses—that is, even if a model correctly guesses an answer through "speculative" way, it should still be penalized. 

Furthermore, according to the research presented by~\citet{razin2025makes}, evaluating reward models based exclusively on accuracy is inadequate; reward variance constitutes a critical component in the RLHF process. The implementation of low confidence penalties will improve the RL performance by introducing variance beyond mere correctness considerations.


Previous research on uncertainty estimation has been conducted~\cite{lin2022teaching,xiong2023can,manakul2023selfcheckgpt}, but most approaches require training an Multi-Layer Perceptron (MLP) as a model probe or accessing internal parameter distribution~\cite{azaria2023internal,burns2022discovering}, leading to issues of poor generalization and high computational demands, making them unsuitable as universal methods for enhancing knowledge reasoning capabilities in RL. Additionally, existing reward models are typically designed for human preference alignment~\cite{wang2024secrets}, concentrating predominantly on instruction adherence and stylistic conformity, while lacking specialized training for STEM domains and mechanisms to penalize low-confidence reasoning.



This paper makes the following key contributions:

1. We introduce the first Correctness and Confidence Reward Model (C2RM) specifically designed for optimizing STEM knowledge capabilities. Unlike traditional reward models that merely collect positive and negative preference pairs based on answer correctness, we additionally gather responses with correct answers but low confidence as negative samples, thereby penalizing responses with low confidence levels. 

2. While technical reports from models such as Qwen3 and Seed1.5-VL~\cite{guo2025seed1} claim to employ reward models for STEM RL training, they disclose no implementation details. We comprehensively reveal the technical specifics of our approach, and open-source our training data, C2RM model checkpoint and policy model checkpoint trained by our C2RM.

3. To validate the effectiveness of our model, we conduct static evaluations (judge bench), inference-time scaling tests (Best-of-N), and post-training RL experiments, complemented by extensive ablation studies for comparative analysis. Experimental evidence demonstrates that our model is comparable to state-of-the-art proprietary models and surpasses mainstream open-source reward models. Furthermore, our ablation studies confirm that integrating both correctness and confidence yields superior results compared to utilizing either factor independently.

\section{Related Work}
\begin{figure*}[t] 
    \centering
        \includegraphics[width=1\textwidth]{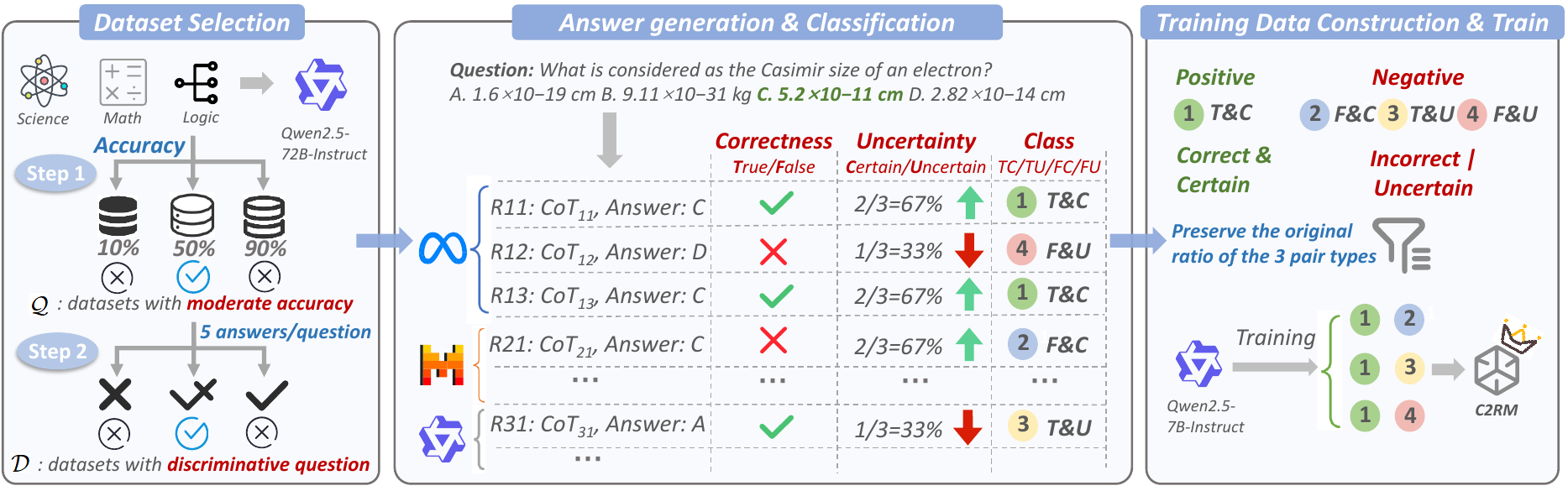}
    \caption{The framework of our reward model data generation and training. We first select high-quality questions where only some of Qwen2.5-72B-Instruct$^\prime$s five answers are correct. Then, for each question, we sample 5 answers (only 3 are shown in the figure for clarity) from each of 3 representative models, label them, and construct training data by treating T\&C as positives and others as negatives. Finally, we train our reward model based on Qwen2.5-7B-Instruct.}
    \label{fig:method}
\end{figure*}

\begin{figure}[t] 
    \centering
        \includegraphics[width=0.5\textwidth]{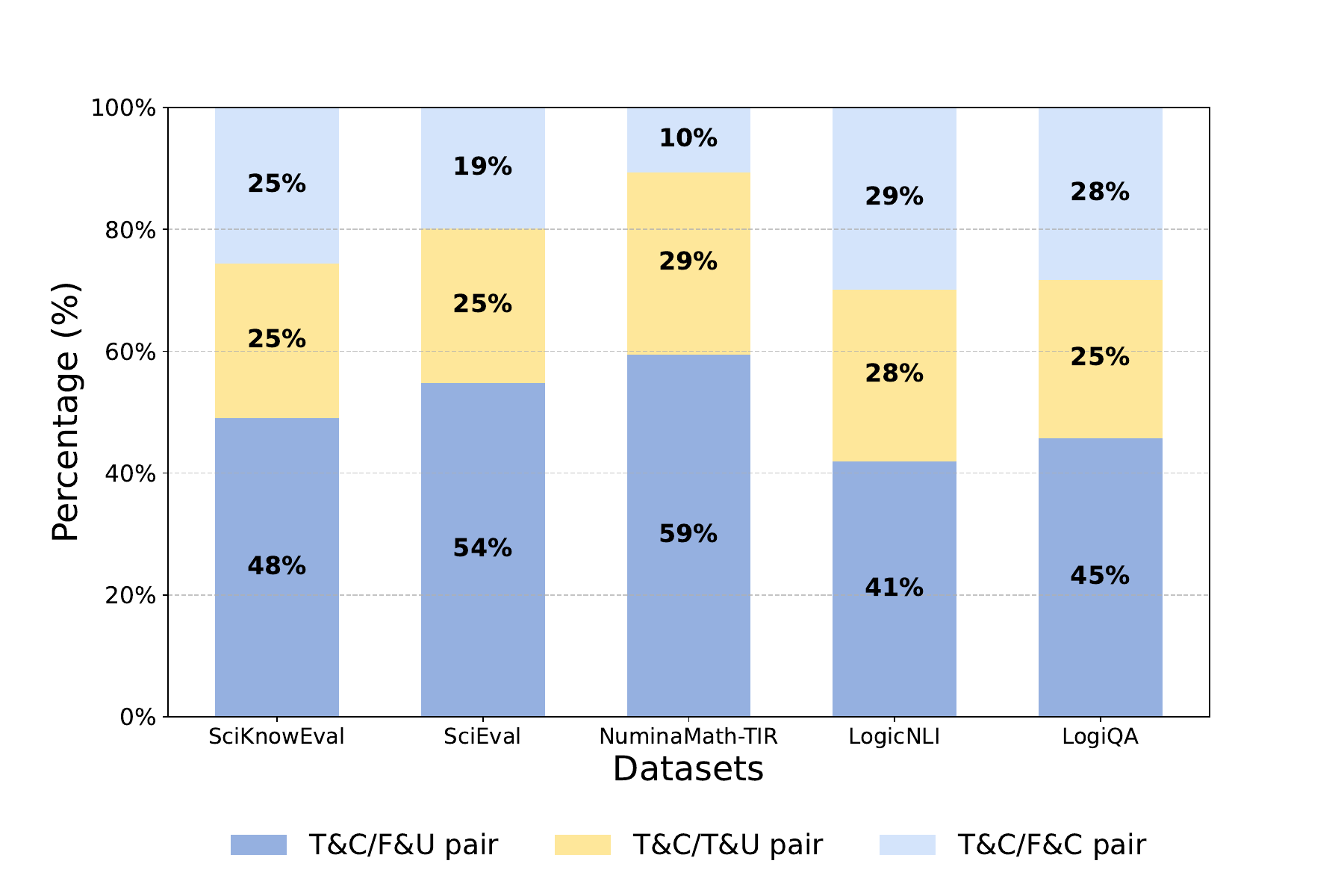}
    \caption{Distribution of different pair types across all datasets. These percentages represent the natural proportions of generated responses, and our training data sampling follows these same natural distributions.}
    \label{fig:pair_proportion}
\end{figure}

\subsection{Uncertainty Estimation}
Methods for measuring uncertainty in LLMs can be classified into four main types~\cite{beigi2024rethinking}. 
Logit-based~\cite{lin2022teaching,mielke2022reducing,kuhn2023semantic} and internal-based~\cite{azaria2023internal,burns2022discovering,li2024inference} methods require access to internal parameters, limiting their applicability in closed-source models. 
In black-box methods, self-evaluation~\cite{kadavath2022language,manakul2023selfcheckgpt,xiong2023can} lets models assess their outputs with confidence prompts. However, its effectiveness is limited by self-awareness, which may lead to overconfidence or inaccuracies~\cite{ji2023towards}. 
Consistency-based methods~\cite{manakul2023selfcheckgpt,wightman2023strength,agrawal2023language} evaluate the agreement among responses, although challenges such as formatting variations and high inference consumption reduce their use and accuracy in real-world applications~\cite{xiong2023can,manakul2023selfcheckgpt}. 
In contrast to these methods, our reward model is designed to implicitly penalize response uncertainty, thereby providing enhanced generalizability, scalability, and efficiency without necessitating access to internal parameters.

\subsection{Reward Model}


Reward models~\cite{liu2024skywork,lou2024uncertainty} are central to LLM's RL training~\cite{wang2024secrets}, with their accuracy and generalization capabilities directly influencing the reinforcement learning outcomes of language models. However, in the era of reasoning models, most reward model research~\cite{liu2024skywork, cai2024internlm2} remains focused on optimizing human preference feedback, without specifically targeting improvements in scientific knowledge reasoning capabilities. We are the first to propose incorporating penalties for policy model uncertainty in STEM RL training, simultaneously considering both correctness and uncertainty during the training process. 

Similar uncertainty concepts have appeared in previous reward model work, such as UP-RLHF~\cite{sun2025uncertainty}, which employs multiple Low-Rank Adaptation (LORA) based reward models and calculates reward values through model ensemble. URM-LLaMa-3.1-8B~\cite{lou2024uncertainty} forces the reward model outputting a normal distribution and combining multiple reward models to estimate reward values by synthesizing multiple normal distributions. Fundamentally, these methods do not directly evaluate policy model uncertainty, but rather employ multiple reward model ensembles to provide greater variance in the RL accuracy training process. 

Unlike the aforementioned methods, our approach generates positive and negative pairs not only from correctness and incorrectness, but also from certain and uncertain answers. We train a Correctness and Confidence Reward Model (C2RM) to evaluate question-answer pairs by punishing their uncertainty, thereby guiding the model to produce higher-quality outputs.

\section{Method}
In this section, we introduce the data construction process and training details of C2RM. The overall framework is illustrated in Fig.~\ref{fig:method}.
\subsection{Dataset Selection}
\label{sec:3.1Dataset Selection}
The training of reward models requires high-quality positive and negative examples. However, many benchmark datasets can not effectively capture the capabilities of modern LLMs. Datasets with either excessively high or low accuracy hinder the construction of meaningful training samples: extremely imbalanced distributions between positive and negative samples can introduce bias into training, while overly easy or overly difficult datasets tend to produce less discriminative outputs, shifting the data distribution and impairing generalization.

To address this, we design a two-step selection process based on the performance of Qwen2.5-72B-Instruct~\cite{qwen2.5}.

\textbf{Step 1:} We first gather a collection of datasets related to STEM. Then we evaluate Qwen2.5-72B-Instruct on each dataset and retain only those where its accuracy falls between 40\% and 70\%. This ensures that the selected datasets are moderately difficult. We denote the filtered datasets as \( \mathcal{Q} \), the candidate question set.

\textbf{Step 2:} For each question \( q \in \mathcal{Q} \), we sample five answers using Qwen2.5-72B-Instruct with a decoding temperature of 0.7. Let \( A_q = \{ a_1, a_2, \ldots, a_5 \} \) denote the set of sampled responses. We then identify a subset of questions where \( A_q \) contains at least one correct and one incorrect answer, indicating inconsistency across outputs. These questions are collected into a new set \( \mathcal{D} \subset \mathcal{Q} \), referred to as the discriminative question set.

This two-step filtering ensures that \( \mathcal{D} \) consists of questions that provoke diverse model behavior, allowing us to extract high-quality contrastive pairs from a shared context. As a result, \( \mathcal{D} \) provides a more informative foundation for training reward models to distinguish correct from incorrect answers.

\paragraph*{Training Datasets.} 
Finally, we select training datasets from three domains:
(1) \textbf{Science}: \textsc{SciKnowEval} (70k multiple-choice questions in biology, chemistry, materials, and physics) and \textsc{SciEval} (18k mostly objective questions across basic science fields);
(2) \textbf{Mathematics}: \textsc{NuminaMath-TIR} (70k numerical-answer math problems targeting symbolic computation and quantitative reasoning);
(3) \textbf{Logical Reasoning}: \textsc{LogicNLI} (20k examples isolating first-order logic from commonsense inference) and \textsc{LogiQA} (8,678 deductive reasoning QA items).

\subsection{Answer generation \& Classification}
\label{sec:3.2Answer generation & Classification}
Following the data selection process in Section~3.1, we construct a seed set of questions with reference answers, denoted as $\mathcal{Q} = \{q_1, q_2, \ldots, q_N\}$, where each $q_i$ has a unique ground truth answer $a_i^{\text{gt}}$.

To ensure diversity in the training data, we use three representative LLMs with varying model sizes and architectures to collect CoT responses: LLaMA-3.1-8B-Instruct~\cite{grattafiori2024llama3herdmodels}, Qwen2.5-72B-Instruct, and Mixtral-8x7B-Instruct-v0.1~\cite{jiang2024mixtralexperts}. Each model generates five responses per question using a decoding temperature of 0.7.

Each response $r_{ij}$ to question $q_i$ is labeled along two dimensions: \textbf{correctness} ($c_{ij} \in \{\text{T}, \text{F}\}$) and \textbf{confidence} ($u_{ij} \in \{\text{C}, \text{U}\}$). A response is marked as \texttt{T} (True) if it exactly matches the ground truth answer; otherwise, it is \texttt{F} (False).

To estimate confidence, we perform $K$ rollouts per question using high-temperature decoding to generate diverse answers $\{a_{i1}, a_{i2}, \ldots, a_{iK}\}$. For any answer $a$, we compute its consistency score:
\[
u_i(a) = \frac{1}{K} \sum_{j=1}^{K} \mathbb{I}(a_{ij} = a)
\]
If $u_i(a) \geq \tau$ (threshold $\tau = 0.5$), the answer is labeled as \texttt{C} (Certain); otherwise, \texttt{U} (Uncertain).

This labeling yields four response types: 
\textit{T\&C} (Correct and Certain), \textit{T\&U} (Correct and Uncertain), \textit{F\&C} (Incorrect and Certain), and \textit{F\&U} (Incorrect and Uncertain). Among them, only \textit{T\&C} responses are used as positive examples for reward model training. The remaining types serve as negative examples to help the model distinguish between preferred and undesired outputs.

Note that a single model may not produce all four types of responses for a given question. For example, if a model generates \textit{T\&C} answers, there may be no \textit{F\&C} examples. To increase data diversity, we use multiple models to answer the same question. Let the model set be $\mathcal{M} = \{M_1, M_2, \ldots, M_L\}$, from which we collect responses to maximize type coverage and enhance reward model supervision.

The collected pool of responses is denoted as:
\[
\mathcal{R} = \left\{
\left.
(q_i,\; r_{ij},\; c_{ij},\; u_{ij},\; m_{ij}) \;\right|
\begin{array}{l}
i = 1, \ldots, N, \\
j = 1, \ldots, K_i
\end{array}
\right\}
\]
where each entry consists of a question $q_i$, a model-generated response $r_{ij}$, its correctness label $c_{ij} \in \{\text{T}, \text{F}\}$, confidence label $u_{ij} \in \{\text{C}, \text{U}\}$, and the model identifier $m_{ij} \in \mathcal{M}$ that produced the response.

\subsection{Training data construction \& Training}

Based on the model responses collected in Sec.~\ref{sec:3.2Answer generation & Classification}, we construct training pairs by comparing different response types. For every question $q_i$, we examine the 15 responses generated by the three models to check whether the following type pairs exist: (1) \textit{T\&C} vs \textit{T\&U} (2) \textit{T\&C} vs \textit{F\&C} (3) \textit{T\&C} vs \textit{F\&U}. If a certain type pair exists for $q_i$, we randomly select one pair of responses from that category to construct a training sample. Therefore, for each question, up to three training samples may be generated, corresponding to each of the three type pairings. The training set can be formally described as:
\[
\mathcal{P} = \left\{
\begin{array}{l}
(q_i,\; r_{ij},\; r_{ik}) \;|\; i = 1,\ldots,N, \\[4pt]
\text{Type}(r_{ij}) = T\&C, \\[4pt]
\text{Type}(r_{ik}) \in \{T\&U, F\&C, F\&U\}
\end{array}
\right\}
\]
\vspace{1em}

Fig.~\ref{fig:pair_proportion} shows the natural distribution of different reasoning pair types across five datasets. Among them, the \textit{T\&C} vs \textit{F\&U} pairs are the most frequent, accounting for roughly half of the samples in each dataset, while the other two pair types appear at similar frequencies. To better reflect real-world data scenarios, we preserve this natural distribution during training.

Considering training efficiency and computational constraints, we select 10K seed questions, resulting in approximately 20K training samples.  Although only one positive-negative pair is generated per question, we randomly sample one type of pair from the different possible combinations for each question while preserving the original distribution of pair types. This approach ensures that each question does not appear too frequently, helping to maintain data diversity and better represent real-world data scenarios.

We use \textbf{Qwen2.5-7B-Instruct}~\cite{qwen2.5} as the base model. The reward model is fine-tuned for 2 epochs with a learning rate of $5 \times 10^{-6}$. Since we use both correctness and confidence as criteria for data construction, we name our reward model \textbf{C2RM} (Correctness and Confidence Reward Model). Details of the data formatting and prompt template are provided in Appendix ~\ref{sec:appendix_1}.

\section{Experiments}
\begin{table*}[t]
\renewcommand{\arraystretch}{1.2}
\newcolumntype{b}{>{\columncolor{blue!4}}c}
\renewcommand{\familydefault}{\rmdefault}
\newlength{\titleshift}
\setlength{\titleshift}{11em} 

\resizebox{\textwidth}{!}{%
\begin{tabular}{l|ccc|ccc|ccc|b}
\toprule
\multirow{2}{*}{\textbf{Method}} & \multicolumn{3}{c|}{\textbf{Llama-3.1-8B-Instruct}} & \multicolumn{3}{c|}{\textbf{Mixtral-8x7B-Instruct-v0.1}} & \multicolumn{3}{c|}{\textbf{Qwen2.5-72B-Instruct}} & \multirow{2}{*}{\textbf{Avg.}} \\
\cmidrule(r){2-4} \cmidrule(r){5-7} \cmidrule(r){8-10}
& GPQA & MATH500 & FOLIO & GPQA & MATH500 & FOLIO & GPQA & MATH500 & FOLIO & \\
\midrule
Pass Avg. & 27.68 & 31.32 & 58.33 & 25.35 & 20.04 & 51.13 & 49.60 & 56.20 & 72.81 & 43.61 \\
\midrule
\multicolumn{11}{c}{\hspace{\titleshift}\textit{Closed-source Models}} \\
\midrule
GPT-4o-2024-11-20 & 35.35 & 37.80 & 64.53 & 28.28 & 26.60 & 57.64 & 56.06 & 56.80 & 74.88 & 48.66 \\
Claude-3-7-sonnet-20250219 & 43.43 & 39.00 & 72.91 & 34.34 & 29.40 & 61.08 & 53.54 & 58.20 & 77.34 & 52.12 \\
Gemini-2.0-flash & 29.80 & 38.00 & 67.98 & 28.28 & 26.40 & 58.62 & 57.07 & 59.20 & 75.37 & 48.97 \\
GPT-4o-mini & 34.34 & 37.40 & 61.08 & 30.81 & 27.80 & 53.20 & 50.51 & 57.40 & 72.41 & 47.22 \\
\midrule
\multicolumn{11}{c}{\hspace{\titleshift}\textit{Open-source Reward Models}} \\
\midrule
URM-LLaMa-3.1-8B & 28.28 & 37.60 & 56.16 & \textbf{30.81} & 26.40 & 53.69 & \underline{53.03} & \underline{58.20} & \underline{71.43} & 46.17 \\
Skywork-Reward-Llama-3.1-8B & \underline{29.29} & \textbf{38.40} & \underline{62.07} & \underline{29.80} & \underline{28.80} & \textbf{56.65} & \textbf{54.55} & 57.80 & 70.44 & \underline{47.53} \\
\textbf{C2RM} & \textbf{29.80} & \underline{38.38} & \textbf{64.04} & 27.27 & \textbf{29.00} & \underline{54.68} & 52.02 & \textbf{59.80} & \textbf{72.91} & \textbf{47.55} \\
\midrule
\multicolumn{11}{c}{\hspace{\titleshift}\textit{Ablation}} \\
\midrule
Correctness-only RM & 29.80 & 39.40 & 59.61 & 30.82 & 28.20 & 60.10 & 47.98 & 60.00 & 70.94 & 47.43 \\
Confidence-only RM & 33.33 & 39.80 & 60.10 & 28.28 & 29.80 & 57.64 & 52.53 & 59.80 & 73.89 & 48.35 \\
\bottomrule
\end{tabular}%
}
\caption{
The overall performance on GPQA, MATH500, and FOLIO across closed-source models, open-source reward models, and our reward model. Our reward model achieves the best performance among open-source models.
}
\label{tab:BoN}
\end{table*}
\begin{table*}[t]
\renewcommand{\arraystretch}{1} 
\newcolumntype{b}{>{\columncolor{blue!4}}c}
\renewcommand{\familydefault}{\rmdefault}
\centering
\resizebox{0.9\textwidth}{!}{
\begin{tabular}{lccccb}
\toprule
\textbf{Models} & \textbf{Knowledge} & \textbf{Math} & \textbf{Reasoning} & \textbf{Coding} & \textbf{Overall} \\
\midrule
\multicolumn{6}{c}{\textit{Closed-source Models}} \\
\midrule
Claude-3-7-sonnet-20250219 & 63.64 & 71.43 & 59.18 & 85.71 & 66.29 \\
GPT-4o-2024-11-20          & 63.64 & 76.79 & 64.29 & 69.05 & 66.57 \\
GPT-4o-mini                & 59.09 & 69.09 & 58.16 & 61.90 & 60.74 \\
Gemini-2.0-flash           & 61.04 & 69.64 & 59.18 & 73.17 & 63.32 \\
\midrule
\multicolumn{6}{c}{\textit{Open-source Reward Models}} \\
\midrule
InternLM2-7B-Reward           & 56.49 & 61.22 & \textbf{71.43} & 50.00 & 59.43 \\
InternLM2-20B-Reward          & \textbf{62.34} & 69.39 & 66.07 & 50.00 & 63.43 \\
Skywork-Reward-Llama-3.1-8B   & 58.44 & \textbf{76.79} & 63.27 & \underline{52.38} & 62.00 \\
URM-LLaMa-3.1-8B              & \textbf{62.34} & \textbf{76.79} & \underline{67.35} & 50.00 & \textbf{64.57} \\
\textbf{C2RM}            & \underline{60.39} & \underline{73.21} & 63.27 & \textbf{69.05} & \underline{64.28} \\
\midrule
\multicolumn{6}{c}{\textit{Ablation}} \\
\midrule
Correctness-only RM & 63.67 & 73.21 & 56.12 & 64.29 & 63.14 \\
Confidence-only RM & 59.09 & 73.21 & 54.08 & 61.90 & 60.29 \\
\bottomrule
\end{tabular}
}
\caption{
The overall performance on JudgeBench across closed-source models, open-source reward models, and our reward model. Our model demonstrates strong overall performance, especially exhibiting remarkable generalization to the unseen coding domain.
}
\label{tab:judgebench-old}
\end{table*}

We conduct a comprehensive evaluation on C2RM from three key perspectives: Best-of-N, static evaluations, and post-training experiments (STEM RL).

\subsection{Baselines}  
(1) \textbf{Closed-source Models}: We evaluate with four widely-used closed-source models, \textit{GPT-4o-2024-11-20}, \textit{GPT-4o-mini}, \textit{Claude-3-7-sonnet-20250219}, and \textit{Gemini-2.0-flash}. The corresponding prompts are detailed in Appendix ~\ref{sec:appendix_2}.
(2) \textbf{Open-source Reward Models}: We include three strong open-source reward models, Skywork-Reward-LLaMA-3.1-8B (Skywork-RM)~\cite{liu2024skywork}, URM-LLaMA-3.1-8B (URM)~\cite{lou2024uncertainty}, and InternLM2-7B-Reward (InternRM)~\cite{cai2024internlm2}. We follow their released scoring methods to evaluate each response.  
(3) \textbf{C2RM and variants}: To evaluate the impact of incorporating confidence in the reward model training data, we design two model variants: ``\textit{Correctness-only RM}'' focuses exclusively on the correctness label, while ``\textit{Confidence-only RM}'' concentrates solely on the confidence label.

All methods adopt a point-wise evaluation protocol. For each question, we select the response with the highest reward score among the five candidates and compute the overall accuracy across the dataset.

\subsection{Best-of-N}
We conduct experiments on three public benchmark datasets: \textsc{GPQA Diamond}~\cite{rein2023gpqagraduatelevelgoogleproofqa}, which focuses on high-quality scientific questions; \textsc{MATH500}~\cite{hendrycks2021measuringmathematicalproblemsolving}, which includes various types of mathematical reasoning problems; and \textsc{FOLIO}~\cite{han2022folio}, which presents challenging reasoning questions. The response generation is performed using the same models as in the training phase. For each question, we generate 5 responses per model with a decoding temperature of 0.7. Note that \textbf{Pass\_Avg}. in Tab~\ref{tab:BoN} represents the average accuracy of the five generated answers for each dataset.

\textbf{Results.} 
Tab.~\ref{tab:BoN} presents the overall performance of BoN test. Compared with existing open-source reward models, our model achieves the best results, demonstrating consistent and comprehensive superiority.
Remarkably, C2RM surpasses GPT-4o-mini , showcasing its powerful capability in evaluating and selecting high-quality responses.
Furthermore, the BoN experiment highlights the practical value of C2RM: it not only selects the correct answer from multiple candidates, but also captures “confidence consistency”, providing a more stable and trustworthy optimization signal.

\subsection{JudgeBench}
We further evaluate the generalization ability of C2RM using the JudgeBench~\cite{judgebench2024}, which is designed to assess LLM-based judges across challenging domains including knowledge, math, reasoning, and coding. To the best of our knowledge, this benchmark represents virtually the only reward model evaluation framework specifically focused on assessing the knowledge reasoning capabilities.

\paragraph*{Results.} 
As shown in Tab.~\ref{tab:judgebench-old}, C2RM achieves strong overall performance with an accuracy of \textbf{64.28\%}, surpassing most mainstream open-source reward models and trailing only URM-LLaMA-3.1-8B by a small margin of 0.29\%. Remarkably, it even outperforms powerful closed-source models such as \texttt{GPT-4o-mini} and \texttt{Gemini-2.0-Flash}.
Notably, C2RM achieves an accuracy of 69.05\% on coding tasks, which were not seen during training. This represents an improvement of \textbf{16.67\%} over Skywork-Reward-LLaMA-3.1-8B and \textbf{19.05\%} over URM-LLaMA-3.1-8B.
These results clearly indicate that C2RM not only possesses strong judgment within trained domains but also generalizes well to novel tasks and fields, effectively distinguishing logically and factually correct answers.

\begin{table*}[t]
\renewcommand{\arraystretch}{1.4} 
\newcolumntype{b}{>{\columncolor{blue!4}}c}
\newcolumntype{d}{>{\columncolor{green!4}}c}
\renewcommand{\familydefault}{\rmdefault}
\centering
\resizebox{0.98\textwidth}{!}{
\begin{tabular}{lccccccccbd}
\toprule
\multirow{2}{*}{\textbf{Reward Models}} & \multicolumn{2}{c}{\textbf{GPQA D.}} & \multicolumn{2}{c}{\textbf{MATH500}} & \multicolumn{2}{c}{\textbf{MMLU-Pro}} & \multicolumn{2}{c}{\textbf{FOLIO}} & \multicolumn{2}{c}{\textbf{Average}} \\
\cmidrule(lr){2-3} \cmidrule(lr){4-5} \cmidrule(lr){6-7} \cmidrule(lr){8-9} \cmidrule(lr){10-11}
 & Acc & Length & Acc & Length & Acc & Length & Acc & Length & Acc & Length \\
\midrule
\multicolumn{11}{c}{\textit{PPO Training}} \\
\midrule
Base & 34.34 & \underline{281.16} & 44.80 & \textbf{343.09} & 49.77 & 132.69 & 58.13 & 23.41 & 46.76 & 195.09 \\
URM-LLaMa-3.1-8B & \textbf{35.86} & 117.76 & 35.60 & 130.91 & 47.16 & 121.09 & 61.58 & 91.25 & 45.05 & 115.25 \\
Skywork-Reward-Llama-3.1-8B & 30.30 & 268.29 & 43.40 & 274.18 & 48.88 & 182.28 & 67.49 & \textbf{191.73} & 47.52 & \underline{228.87} \\
internlm2-7b-reward & 28.79 & 184.21 & 43.80 & 219.33 & 46.05 & \underline{186.61} & 61.08 & 142.57 & 44.93 & 183.18 \\
Rule-based & 28.79 & 219.41 & \textbf{48.80} & \underline{289.74} & \underline{54.19} & 184.07 & \underline{67.98} & 137.94 & \underline{49.94} & 207.79 \\
\textbf{C2RM} & \underline{35.35} & \textbf{302.50} & \underline{47.80} & 252.84 & \textbf{55.85} & \textbf{234.55} & \textbf{73.40} & \underline{190.13} & \textbf{53.10} & \textbf{245.01} \\
\midrule
\multicolumn{11}{c}{\textit{Ablation}} \\
\midrule
Correctness-only RM & 29.80 & 195.86 & 51.00 & 228.87 & 52.52 & 132.38 & 61.08 & 125.40 & 48.60 & 170.63 \\
Confidence-only RM & 32.32 & 179.43 & 48.2 & 206.20 & 53.07 & 128.00 & 57.64 & 91.93 & 47.81 & 151.39\\
\bottomrule
\end{tabular}
}
\caption{
The overall performance of policy models trained with different reward models across multiple benchmarks including GPQA Diamond, MATH500, MMLU-Pro, and FOLIO. C2RM demonstrates the strongest overall performance with an average score of 53.10\%, particularly excelling in MMLU-Pro and FOLIO benchmarks.
}
\label{tab:reward-models-comparison}
\end{table*}
\subsection{RL}
To validate the effectiveness of our reward model in reinforcement learning, we apply the PPO algorithm for training and evaluation in STEM scenarios.

\textbf{Proximal Policy Optimization (PPO)} is an on-policy reinforcement learning algorithm designed to improve a stochastic policy \(\pi_\theta\) through iterative updates that balance progress and stability. At each time step \(t\), we compute the probability ratio
\[
r_t = \frac{\pi_\theta(a_t \mid s_t)}{\pi_{\theta_{\mathrm{old}}}(a_t \mid s_t)}\,
\]
which measures the change in action probability under the updated policy relative to the previous one.
The PPO loss $\mathcal{L}_{\text{PPO}}$ is computed based on the advantage estimates $A$, which measure how much better an action is compared to the average action in a given state:
\begin{equation*}
\small
    \begin{split}
        \mathcal{L}_{\mathrm{PPO}}(\theta) 
        &= \min \Biggl( 
            \frac{\pi_\theta}{\pi_{\mathrm{ref}}} \cdot A, \;
            \mathrm{clip}\left( 
                \frac{\pi_\theta}{\pi_{\mathrm{ref}}}, 
                1 - \epsilon, 
                1 + \epsilon 
            \right) \cdot A 
        \Biggr)
    \end{split}
\end{equation*}
where $\pi_{\text{ref}}$ represents the reference policy (typically from the previous iteration), and $\epsilon$ is a hyperparameter that controls the clipping range (usually set to 0.2). 

\textbf{Settings.}
We sampled 30K question from dataset $\mathcal{D}$ mentioned in section~\ref{sec:3.1Dataset Selection} in RL training.
Qwen2.5-7B-Instruct is used as the policy model to be optimized. To evaluate the effectiveness of C2RM, we compare it against the following baselines: (1) \textbf{Base}, which refers to the original, untrained Qwen2.5-7B-Instruct model; (2) \textbf{Open-source reward models}, including three widely-used models: \textit{URM-LLaMa-3.1-8B}, \textit{Skywork-Reward-LLaMA-3.1-8B}, and \textit{InternLM2-7B-Reward}; (3) \textbf{Rule-based}, which does not rely on a reward model, but instead assigns rewards solely based on whether the model's final answer matches the ground truth, ignoring the CoT process.

For evaluation, we adopt four benchmark datasets: \textsc{GPQA}, \textsc{MATH500}, \textsc{FOLIO}, and \textsc{MMLU-Pro}, which represent scientific knowledge, mathematics, logical reasoning, and multidisciplinary understanding, respectively. These benchmarks allow for a comprehensive assessment of the impact of different training strategies on the policy model's performance.

\textbf{Results.}
Based on the results in Sec.~\ref{sec:4.3.1Training dataset selection}, we confirm that \( \mathcal{D} \) is more suitable for PPO training. Therefore, we perform further refined PPO experiments using the same 30K samples from \( \mathcal{D} \), evaluating the resulting policy models in terms of their accuracy on the test sets and the average number of tokens in their responses.

Our experimental results demonstrate the outstanding effectiveness of C2RM as a reward model for reinforcement learning in STEM tasks. As shown in Table~\ref{tab:reward-models-comparison}, C2RM consistently outperforms all baselines across multiple benchmarks, achieving an average accuracy of 53.10\%, outperform 6.34\% over the base model and achieve 6\% to 8\% higher than other open‑source reward models. Notably, C2RM achieves an impressive \textbf{73.40\%} accuracy on \textit{FOLIO}, outperforming the second-best model (rule-based, 67.98\%) by 5.42\% and the base model by 15.27\%. These significant gains underscore C2RM’s ability to guide policy models through complex logical reasoning tasks.

Beyond raw accuracy, C2RM also promotes the generation of more comprehensive and detailed responses, as evidenced by consistently longer average token lengths across benchmarks. Under C2RM’s guidance, the average response length reaches 245.01 tokens-substantially exceeding both the rule-based model (207.79) and Skywork-RM (228.87). This indicates that C2RM not only rewards correctness but also encourages completeness of reasoning, a critical factor in STEM problem solving where the reasoning process is as important as the final answer. 

Notably, URM, which demonstrated excellent performance on the JudgeBench in Tab.~\ref{tab:judgebench-old}, exhibited a significant reduction in response length, compressing from an average of 195 tokens in the original base model to 115 tokens. This optimization actually contradicts the goal of enhancing the model's reasoning capabilities, resulting in the lowest  optimized policy model performance among all baselines.

Detailed reward and response length curves, as well as case studies comparing policy model responses before and after PPO training, can be found in Appendix~\ref{sec:RL Training Curves}.

\subsection{Ablation Study}
To evaluate the impact of incorporating confidence as an additional dimension in the training data on reward model performance, we design two alternative data construction schemes and conduct a comparative experiment. Specifically, based on the answer generation process described in Sec.~\ref{sec:3.2Answer generation & Classification}, we construct two types of training data from the five generated responses per question: (1) \textbf{Correctness-only}: a response is labeled as positive if its final answer matches the ground-truth answer; otherwise, it is labeled as negative, regardless of the model's confidence; (2) \textbf{Confidence-only}: a response is labeled as positive if its predicted confidence exceeds a predefined threshold of 50\%; otherwise, it is labeled as negative, regardless of answer correctness.

Under both construction strategies, we use 10k seed questions, sampling one positive and one negative example per question, resulting in a total of 20k training samples. All training hyperparameters are kept consistent with those used in the main experiments.

We evaluate the two trained models using Best-of-N (BoN), JudgeBench, and PPO RL. The corresponding results are reported in the ablation sections of Tables~\ref{tab:BoN}, \ref{tab:judgebench-old}, and \ref{tab:reward-models-comparison}.

In BoN experiment, the overall performance of the \textit{Confidence-only RM} achieves a surprisingly strong average accuracy of 48.35\%, achieving comparable performance to GPT-4o-2024-11-20 and Gemini-2.0-flash. This result underscores the critical role of confidence information in reward modeling: even when correctness is ignored, leveraging confidence alone can lead to competitive performance. 

This may be attributed to the fact that the training data synthesis for this paper originated from these models. The \textit{Confidence-only RM} has already learned to recognize certain linguistic patterns related to uncertainty from these models and uses these patterns to evaluate quality. In the BoN test, this approach might be more effective than relying on the objective correctness of answers.

In JudgeBench and RL experiments, both ablation models perform worse than C2RM, highlighting their limited generalization capability in more complex evaluation settings. Particularly in the response length test during RL, we observed that both variants led to a decrease in response length. 

One of the primary reasons for C2RM's superior performance, as mentioned in the introduction section, is its ability to penalize correct responses generated through low-confidence reasoning processes, thereby enhancing the quality of cognitive deliberation. Another possible explanation, as mentioned in the introduction, is that low confidence penalties introduce greater reward variance, with both mean and variance being crucial elements for optimization algorithms.

In summary, relying solely on either correctness or confidence leads to suboptimal model performance. Combining both dimensions in data construction is essential for achieving the best results, thereby strongly validating the effectiveness of our proposed method.

\section{Conclusion}
In this work, we present a correctness and confidence reward modeling approach (C2RM) to enhance the reasoning capabilities of small-scale language models in STEM domains. By explicitly penalizing low-confidence responses—even when the final answer is correct, our method addresses the limitations of rule-based RL, which often rewards accidental or speculative correct answers. We release full implementation details and demonstrate that our reward model effectively guides LLM toward more reliable and interpretable reasoning behavior, without relying on huge-scale foundation models. Our results highlight a scalable and accessible path for knowledge-intensive post-training under resource constraints. 

\clearpage
\section{Limitations}


Our model currently does not support multimodal data. We plan to expand this capability by collecting multimodal data in the future. We have only trained a 7B model, without providing smaller or larger variants, because the 7B size aligns with mainstream reward model. We selected only five open-source datasets as our training data sources. While we believe that expanding data sources and enhancing data diversity would improve our model's capabilities, we opted for only five data sources to demonstrate the effectiveness of our approach. Regarding data volume, we selected only 10K positive-negative pairs for RM training, whereas other baselines, such as URM and Skywork-RM, typically utilize approximately 80K data.

\bibliography{custom}

\clearpage

\appendix

\section{Appendix}
\label{sec:appendix}

\subsection{Details of C2RM Training Data \& Settings} 
\label{sec:appendix_1}
\subsubsection{Model Answer Generation}
In Sec.~\ref{sec:3.2Answer generation & Classification}, we describe how we generate responses based on the filtered dataset. Specifically, to obtain model outputs with Chain-of-Thought (CoT) reasoning and to facilitate the extraction of final answers, we use the prompt shown in Tab.\ref{tab:100model_prompt} to guide the model's generation. It is important to note that the CoT process plays a crucial role in our data collection. We expect the reward model to capture the differences in CoT between positive and negative samples, thereby improving its ability to identify correct and confident responses.
\begin{table}[htbp]
\centering
\renewcommand{\arraystretch}{1.5}
    \begin{tabularx}{0.45\textwidth}{X}
    \toprule
    Read the question, analyze step by step and provide your answer. Use the following format to answer:
    ```Explanation: [insert step-by-step analysis here]
    Answer: [ONLY the final answer; not a complete sentence]
    ```
    Please make sure to analyze step by step before giving the answer.
    Only give me the reply according to this format, don't give me any other words. \\
    \bottomrule
    \end{tabularx}
  \caption{
    The prompt used for model answer generation.
  }
  \label{tab:100model_prompt}
\end{table}

\subsubsection{Details of Reward Model Training}
We train the reward model using supervised fine-tuning. The goal is to enable the model to identify high-quality responses that are both \textbf{correct} and \textbf{certain}, and assign them higher scores.

Each training example is formatted using an instruction-tuning style, consisting of an \texttt{instruction}, an \texttt{input}, and an \texttt{output} (\texttt{"Yes"} or \texttt{"No"}). Specifically, responses labeled as both correct and certain are assigned \texttt{"Yes"}, while all other types are labeled as \texttt{"No"}. 

\begin{figure*}[!ht]
\begin{lstlisting}
{
    "instruction": "Given the following Question and the corresponding Answer provided by a model, you are required to assess whether the model is certain about its answer. If the model is certain about its answer, output 'Yes'. If the model is uncertain about its answer, output 'No'.",
    "input": "Question:\n[question]\n\nModel's Answer:\n[answer]",
    "output": "Yes/No"
}
\end{lstlisting}
\caption{Prompt for closed-source model test in BoN.}
\label{lst:certainty_assessment}
\end{figure*}

After training, we evaluate the reward model's ability to score model responses using the predicted probability of the token \texttt{"Yes"}. Under a constrained decoding setup where the output vocabulary is limited to $\mathcal{V}_{\text{allowed}} = \{\text{"Yes"}, \text{"No"}\}$, the reward score is defined as the probability of generating \texttt{"Yes"} given the input:
\[R(x) = P_\theta(y = \text{"Yes"} \mid x)\]
where $x$ denotes the input (including the question and the model's answer), and $P_\theta$ represents the model's output distribution. Let the target sequence for \texttt{"Yes"} consist of tokens $\{t_1, t_2, \ldots, t_n\}$. We compute the log probability of the sequence as:
\[
\log P_\theta(y = \text{"Yes"}) = \sum_{i=1}^{n} \log P_\theta(t_i \mid x, t_{<i})
\]
Then, the final reward score is given by:
\[
R(x) = \exp\left( \sum_{i=1}^{n} \log P_\theta(t_i \mid x, t_{<i}) \right)
\]
This scalar score serves as a quantitative indicator of the model’s confidence in the response and can be directly used in preference modeling or RLHF pipelines.

\subsection{Closed source model BoN testing method} 
\label{sec:appendix_2}
To evaluate the best-of-n (BoN) capability of closed-source models, we adopt the prompt shown in Tab. \ref{tab:100close_llm_bon_prompt}, which is designed to assign pointwise scores to the responses of each question. Among the five candidate answers, the one with the highest score is selected, and its correctness is used to compute the final accuracy.
\begin{table}[t]
\begin{flushright}  
\centering
\renewcommand{\arraystretch}{1.5}
    \begin{tabularx}{0.9\columnwidth}{X}  
    \toprule
    <Instruction> Given the following Question and the corresponding Answer provided by a model, assess the probability that the answer is correct. Please provide only a single number between 0 and 1 (not inclusive) representing the probability, where 0 means completely incorrect and 1 means completely correct. Your answer should be formatted as:
    ```json
    {{"probability": 0.5}}
    ```
    reason and analysis: <your analysis>
    </Instruction>\\
    <Question and the answer need to be scored>: \\
    Question: \{question\} \\
    Model's Response:\\
    \{model response\} \\
    \bottomrule
    \end{tabularx}
  \caption{
    The prompt used for testing closed-source models in the BoN task.
  }
  \label{tab:100close_llm_bon_prompt}
\end{flushright}
\end{table}

\begin{figure}[t] 
    \centering
        \includegraphics[width=0.48\textwidth]{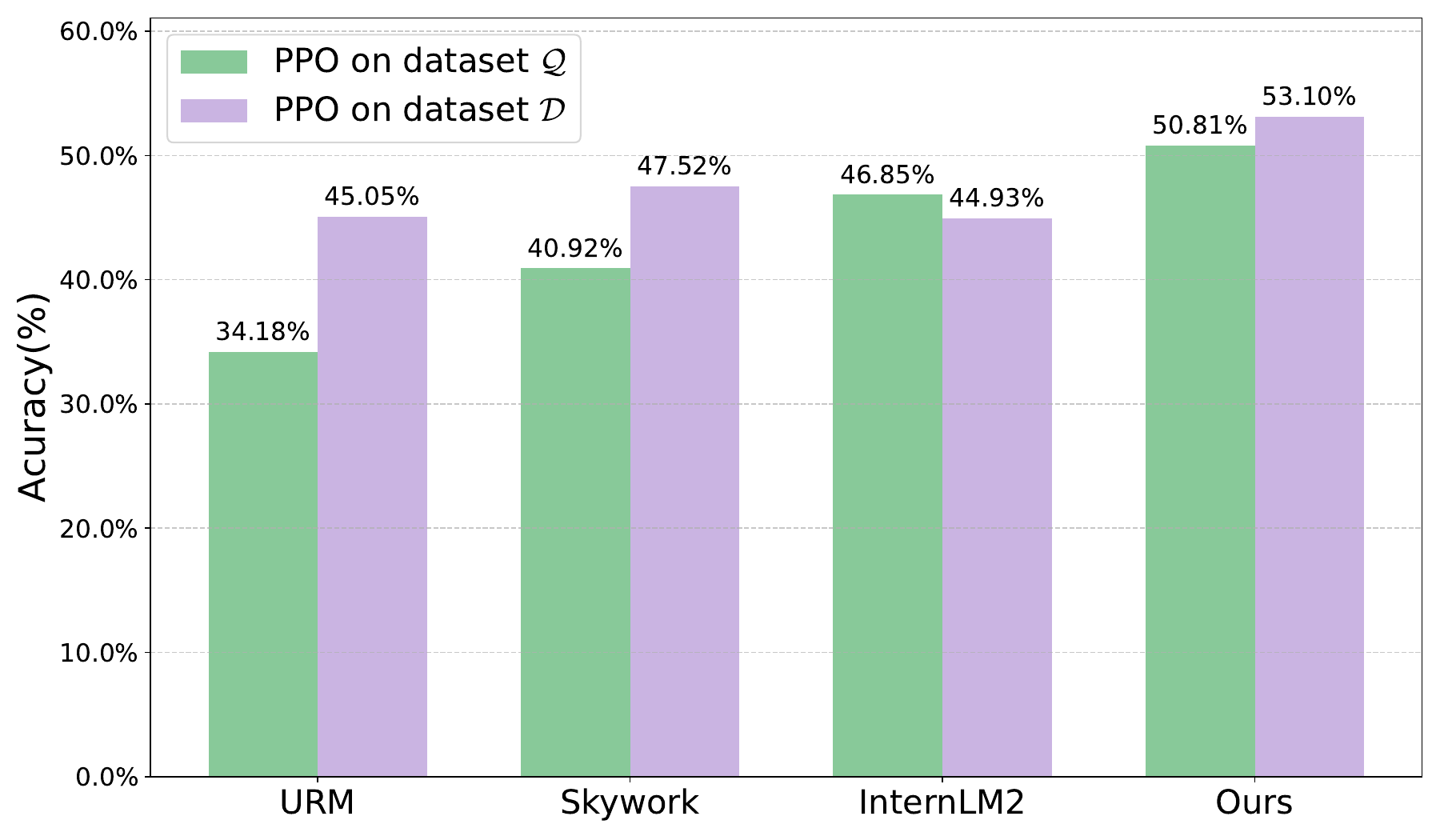}
    \caption{Average accuracy on GPQA-Diamond, MATH500, FOLIO, and MMLU-Pro after PPO training using 30k samples from $\mathcal{Q}$ and $\mathcal{D}$, with different reward models respectively. }
    \label{fig:bigq_smallq}
\end{figure}

\subsection{RL Training dataset selection}
\label{sec:4.3.1Training dataset selection}
Before initiating formal PPO training, it is crucial to determine which type of data is most effective for optimizing the policy model. Based on the data selection process detailed in Sec.~\ref{sec:3.1Dataset Selection}, we construct two datasets, \( \mathcal{Q} \) and \( \mathcal{D} \), each reflecting different selection criteria. We then evaluate the impact of these datasets on PPO using distinct reward models.

Specifically, we sample 30k seed questions from each of \( \mathcal{Q} \) and \( \mathcal{D} \) as the PPO training sets, train the policy models for 3 epochs, and evaluate the resulting models on the aforementioned test sets.

\textbf{Results.}
As shown in Fig.~\ref{fig:bigq_smallq}, PPO training on dataset $\mathcal{D}$ outperforms that on $\mathcal{Q}$. This result highlights the advantage of our two-step selection strategy: while $\mathcal{Q}$ ensures moderate difficulty, $\mathcal{D}$ further offers more informative and discriminative training signals. Consequently, $\mathcal{D}$ serves as a better training set for policy model optimization.

\subsection{RL Training Curves}
\label{sec:RL Training Curves}
\begin{figure*}[t] 
    \centering
        \includegraphics[width=0.9\textwidth]{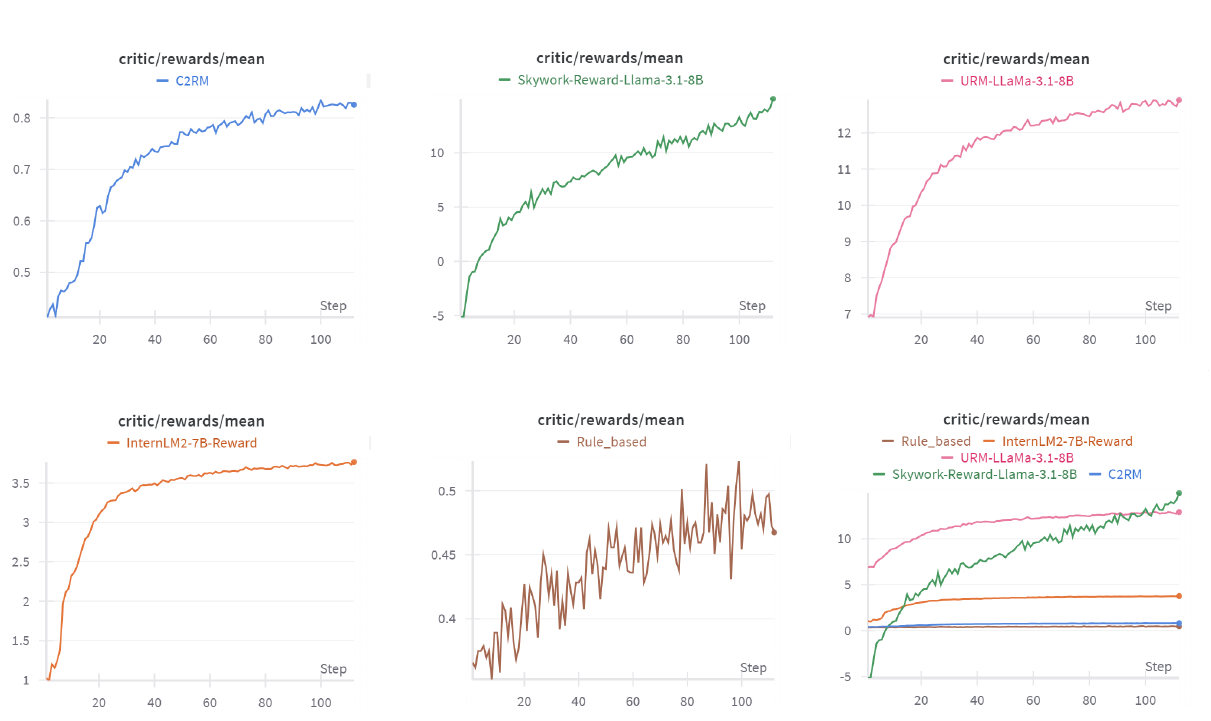}
    \caption{Reward curve of the reward model during PPO training.}
    \label{fig:reward}
\end{figure*}
\begin{figure*}[t] 
    \centering
        \includegraphics[width=0.9\textwidth]{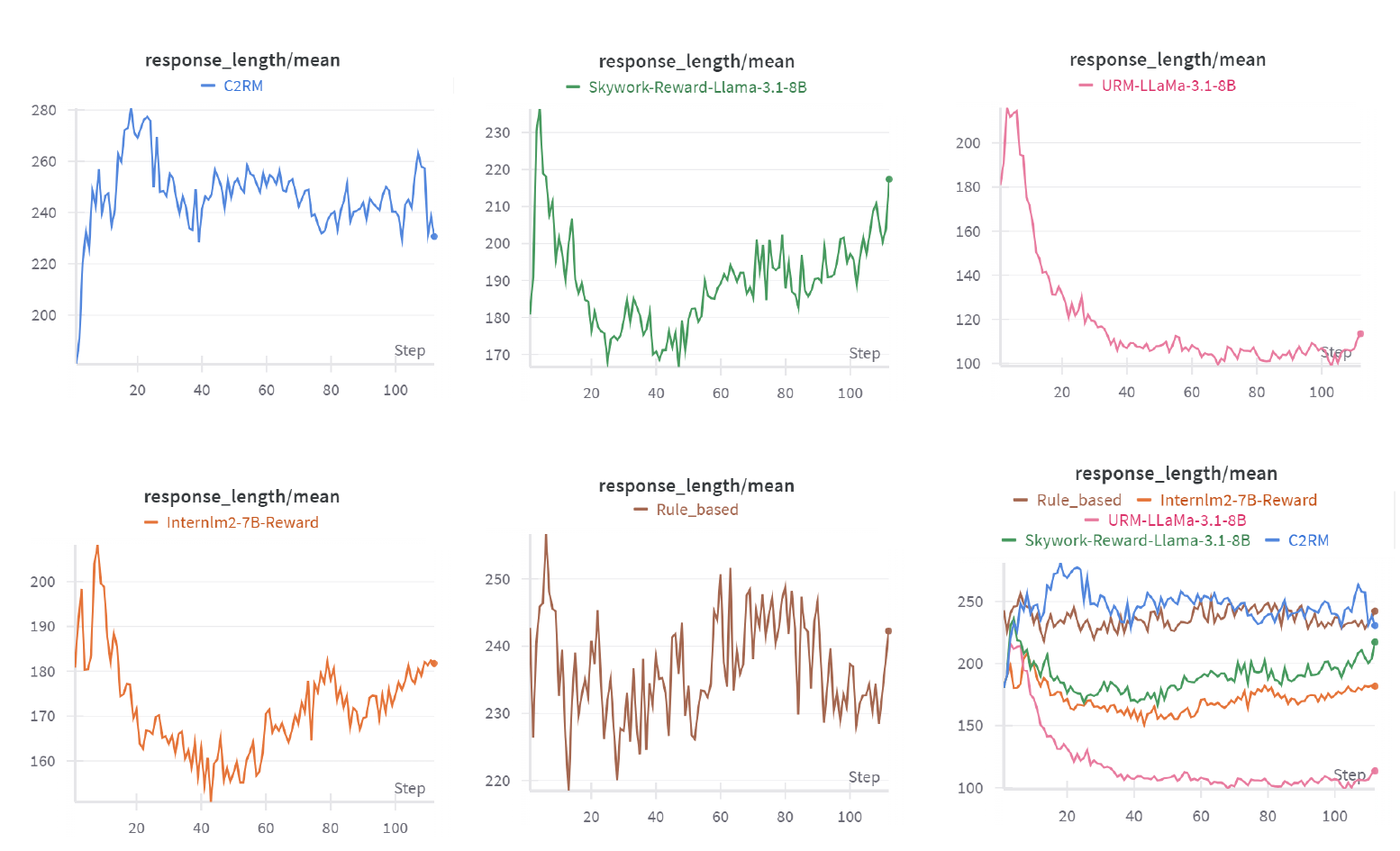}
    \caption{The response length curve of the reward model during PPO training.}
    \label{fig:length}
\end{figure*}

\subsubsection{Reward Curves}
Analyzing the reward curves from different models during PPO training reveals a general upward trend across all models, indicating effective policy optimization. Notably, our C2RM model demonstrates particularly impressive performance characteristics compared to other reward models. When examining its trajectory, we observe a rapid initial increase followed by sustained improvement throughout the training process, eventually stabilizing at a high reward level. It's important to recognize that the absolute reward values between different reward models cannot be directly compared due to their varying normalization approaches and scoring ranges. While some reward models operate on unbounded scales, our C2RM model specifically constrains rewards between 0 and 1, which makes its performance trajectory appear less dramatic when viewed alongside other models with wider output ranges. Despite this normalization difference, C2RM's consistent upward trend and eventual stabilization above 0.8 in its normalized range represents exceptional performance, especially considering that approaching the upper bound of 1.0 in our tightly constrained reward space indicates near-optimal alignment with desired response characteristics. This robust improvement pattern suggests that C2RM provides a stable and effective reward signal for guiding language model alignment throughout the training process.

\subsubsection{Response Length Curves}
The average response length metrics during PPO
The average response length metrics during PPO training provide compelling evidence of our C2RM reward model's exceptional performance compared to alternative approaches. When examining the evolution of response length across different reward models, C2RM stands out remarkably by maintaining and even increasing the response length throughout the training process, stabilizing at approximately 250 tokens. This represents a significant achievement in reinforcement learning for language models, where response length preservation is notoriously challenging.

In stark contrast, other prominent reward models exhibit concerning degradation patterns: URM-LLaMa-3.1-8B shows dramatic length collapse, plummeting from over 200 tokens to merely 100 tokens by training completion—a catastrophic 50\% reduction that severely compromises the model's ability to provide comprehensive reasoning. Similarly, Internlm2-7B-Reward experiences substantial degradation, dropping below 160 tokens before slightly recovering, while Skywork-Reward-LLaMa-3.1-8B initially collapses from 230 tokens to 170 tokens before its partial recovery. This comparison highlights C2RM's unique capability to incentivize detailed, thorough responses without the length collapse phenomenon that plagues competing approaches. 

Response length serves as a crucial proxy for reasoning quality, as shorter responses typically lack the step-by-step reasoning and nuanced explanations necessary for complex problem-solving. Our C2RM model's ability to maintain robust response lengths indicates its sophisticated alignment with desirable response characteristics, making it particularly well-suited for applications requiring detailed reasoning processes or complex explanations. This distinctive performance advantage demonstrates that C2RM provides a more balanced and holistic reward signal that successfully encourages both high-quality content and appropriate response thoroughness.

\subsubsection{Case Presentation}
\begin{table*}[]
\small
    \begin{tabularx}{\linewidth}{X}
    \toprule
    \color{gray}{/* \textit{Astronomy problem} */}\\
    Astronomers are observing a planet with a radius of 1 Earth radius in a circular orbit, with a transit impact parameter of 0.2, around a star that has a radius 1.5 times that of the Sun. This planet has an orbital period of 3 days. What is the maximum orbital period of a second planet, with a radius of 2.5 times that of Earth, in the same system that will exhibit both transit and occultation events? Assume that the second planet also has a circular orbit and shares the same orbital plane as the previously mentioned transiting planet.
    
    \textbf{Options:}
    A. $\sim$ 12.5 days
    B. $\sim$ 33.5 days
    C. $\sim$ 7.5 days
    D. $\sim$ 37.5 days
    
    \textbf{Correct Answer:} B
    \\
    \color{gray}{/* \textit{Model output before training} */}\\
    \textcolor{red}{Explanation:} 
    1. The transit impact parameter (b) is 0.2, which means the second planet can transit the star.
    2. The orbital period of the first planet is 3 days. Using Kepler's third law, \(T^2 \propto R^3\), where \(T\) is the orbital period and \(R\) is the semi-major axis of the orbit.
    3. The radius of the second planet is 2.5 times that of Earth, but this does not affect the orbital period calculation for the second planet.
    4. To find the maximum orbital period of the second planet that will exhibit both transit and occultation, we use the relationship \(T_2^2 / T_1^2 = R_2^3 / R_1^3\). Given \(T_1 = 3\) days, \(R_2 = 2.5 R_1\), and \(R_1\) is the radius of the first planet's orbit.
    5. Plugging in the values, we get \(T_2^2 / 3^2 = (2.5)^3 / 1^3\).
    6. Solving for \(T_2\), we get \(T_2 = 3 \times (2.5)^{3/2} \approx 3 \times 3.95 \approx 11.85\) days, which is closest to 12.5 days.
    
    Answer: \textcolor{red}{A}
    \\
    \color{gray}{/* \textit{Model output after training} */}\\
    \textcolor{green}{Explanation:} To determine the maximum orbital period of the second planet that will exhibit both transit and occultation events, we need to consider the geometry and dynamics of the system. The transit impact parameter (b) is given as 0.2, which means the orbit of the first planet is slightly inclined relative to the line of sight. For the second planet to exhibit both transit and occultation, its orbit must be in a specific configuration relative to the first planet's orbit. The key factor is the orbital period ratio, which is related to the semi-major axes of the orbits. Given the first planet's orbital period of 3 days, we can use Kepler's third law to find the relationship between the periods of the two planets. However, the exact calculation involves complex orbital mechanics and the specific geometry of the system. Given the options, the most plausible answer based on the provided information and the need for both transit and occultation is closest to 33.5 days. This value balances the need for the second planet to have a significant enough period to allow for both events while fitting within the provided options.
    
    Answer: \textcolor{green}{B}
    \\
    \bottomrule
    \end{tabularx}
  \caption{
    Comparison of policy model before and after PPO training on a scientific reasoning problem.
  }
  \label{tab:astronomy_reasoning_improvement}
\end{table*}
\begin{table*}[]
\small
    \begin{tabularx}{\linewidth}{X}
    \toprule
    \color{gray}{/* \textit{Mathematical problem} */}\\
    Jim and Martha are standing together at the corner of a rectangular field. Jim walks diagonally across the field. Martha gets to the same location by walking along its length and width. The field is 300 feet wide and 400 feet long. How many feet less than Martha does Jim walk?
    \\
    \color{gray}{/* \textit{Standard solution} */}\\
    The diagonal of the field is $\sqrt{300^2+400^2}=500$ feet long so Jim walks 500 feet. Two adjacent sides of the field are $300+400=700$ feet long so Martha walks 700 feet. Jim walks $700-500=\boxed{200}$ feet less than Martha.
    \\
    \color{gray}{/* \textit{Model output before training} */}\\
    \textcolor{red}{Explanation:} 
    1. Calculate the diagonal distance Jim walks using the Pythagorean theorem: \( \sqrt{300^2 + 400^2} \).
    2. Calculate the total distance Martha walks, which is the sum of the length and width: \( 300 + 400 \).
    3. Subtract the distance Jim walks from the distance Martha walks to find the difference.
    
    Answer: \textcolor{red}{\boxed{100}}
    \\
    \color{gray}{/* \textit{Model output after training} */}\\
    \textcolor{green}{Explanation:} Let's first calculate the distance Jim walks. Since Jim walks diagonally across the rectangular field, we can use the Pythagorean theorem to find the length of the diagonal. The width and length of the field are the two legs of a right triangle, so the diagonal \(d\) is given by:
    \[ d = \sqrt{300^2 + 400^2} = \sqrt{90000 + 160000} = \sqrt{250000} = 500 \text{ feet} \]
    
    Next, we calculate the distance Martha walks. Since Martha walks along the length and width of the field, she walks:
    \[ 300 + 400 = 700 \text{ feet} \]
    
    The difference in the distances walked by Martha and Jim is:
    \[ 700 - 500 = 200 \text{ feet} \]
    
    Therefore, Jim walks 200 feet less than Martha.
    
    Answer: \textcolor{green}{\boxed{200}}
    \\
    \bottomrule
    \end{tabularx}
  \caption{
    Comparison of policy model before and after PPO training on a mathematical reasoning problem.
  }
  \label{tab:math_reasoning_improvement}
\end{table*}
\begin{table*}[]
\small
    \begin{tabularx}{\linewidth}{X}
    \toprule
    \color{gray}{/* \textit{Logical reasoning problem} */}\\
    Please determine whether the conclusion is true, false, or uncertain based on these premises.\\
    \textbf{Premises:}\\
    1. All employees who schedule a meeting with their customers will go to the company building today. \\
    2. Everyone who has lunch in the company building schedules meetings with their customers. \\
    3. Employees have lunch either in the company building or at home.\\
    4. If an employee has lunch at home, they are working remotely from home.\\
    5. All employees who are in other countries work remotely from home. \\
    6. No managers work remotely from home. \\
    7. James will appear in the company today if and only if he is a manager. \\
    \textbf{Conclusion:} James has lunch in the company.\\
    \textbf{Expected Answer:} Uncertain
    \\
    \color{gray}{/* \textit{Model output before training} */}\\
    \textcolor{red}{Answer: False}
    \\
    \color{gray}{/* \textit{Model output after training} */}\\
    \textcolor{green}{Explanation:} Let's analyze the premises step by step:
    1. All employees who schedule a meeting with their customers will go to the company building today.
    2. Everyone who has lunch in the company building schedules meetings with their customers.
    3. Employees have lunch either in the company building or at home.
    4. If an employee has lunch at home, they are working remotely from home.
    5. All employees who are in other countries work remotely from home.
    6. No managers work remotely from home.
    7. James will appear in the company today if and only if he is a manager.
    
    From these premises, we can deduce the following:
    - If James is a manager, he will appear in the company today.
    - Since no managers work remotely from home, James cannot be working remotely from home.
    - Therefore, James must have lunch in the company building to appear in the company today.
    - If James has lunch in the company building, he schedules meetings with his customers.
    - If he schedules meetings with his customers, he will go to the company building today.
    
    Given the conclusion that "James has lunch in the company," we need to check if this is necessarily true based on the premises. The premises do not provide enough information to definitively conclude that James must have lunch in the company. It is possible that James could be in a different country and working remotely, or he could be a non-manager who does not need to schedule meetings. Therefore, the conclusion is not necessarily true based on the given premises.
    
    Answer: \textcolor{green}{Uncertain}
    \\
    \bottomrule
    \end{tabularx}
  \caption{
    Comparison of policy model before and after PPO training on a logical reasoning problem.
  }
  \label{tab:logic_reasoning_improvement}
\end{table*}
To demonstrate the effectiveness of using C2RM as the reward model for PPO training on Qwen2.5-7B-Instruct, we present one case each from the domains of science, mathematics, and logic in Fig.~\ref{tab:astronomy_reasoning_improvement},Fig.~\ref{tab:math_reasoning_improvement} and Fig.~\ref{tab:logic_reasoning_improvement}. As shown, the original model produces incorrect answers with short and incomplete reasoning. After PPO training, however, the model generates correct answers along with more complete reasoning paths.

\end{document}